# An Ensemble Deep Learning Approach for Reliable and Scalable Lemon Leaf Disease Classification

Shayan Abrar
*Dept. of CSE*
*American International University-Bangladesh*
Dhaka, Bangladesh
shayanabrar7@gmail.com

Sudeepta Mandal
*Dept. of CSE*
*East West University*
Dhaka, Bangladesh
msudeep996@gmail.com

Abdul Awal Yasir
*Dept. of CSE*
*North South University*
Dhaka, Bangladesh
yasir.nafis201@gmail.com

Sonjoy Bhattacharjee
*Dept. of CSE*
*East West University*
Dhaka, Bangladesh
sonjoybhattacharjee159@gmail.com

Sadman Haque Bhuiyan
*Dept. of CSE*
*East West University*
Dhaka, Bangladesh
alifsadman798@gmail.com

Samanta Ghosh
*Dept. of CSE*
*East West University*
Dhaka, Bangladesh
Samantaewu28@gmail.com

Rafi Ahamed
*Dept. of CSE*
*East West University*
Dhaka, Bangladesh
fidaahamed15@gmail.com

***Abstract*— Early detection of plant diseases is crucial to plants and for the farmers. Plant diseases reduce fruit yield and quality, and plants are more susceptible to other stresses when they are infected. The lemon leaf disease dataset contains 1354 images. The dataset has 9 classes. Among the 9 classes only one class is for healthy leaf, and the other 8 classes are leaf diseases. The dataset was split into training (70%), testing (15%) and validation (15%) sets after comprehensive preprocessing. Two pretrained models (InceptionV3 and MobileNetV2) were applied and then combined these models using an ensemble technique to boost robustness. Ensemble models showed a promising performance of 99.27% accuracy. Adversarial Training is applied to improve models' ability and ensure reliable predictions under noisy data. Grad-CAM visualization highlights the important regions of leaf images that validate the model prediction with confidence level.**

## I. Introduction

Lemon is one of the most important citrus fruits in Bangladesh and all over the world. This fruit plays a crucial role in the rural economy. Because of its nutritional and medical value, the demand for this fruit is always high. Lemon plants are vulnerable to many leaf diseases like citrus canker, anthracnose, greasy spot, and sooty mold. These diseases affect the fruit quality and quantity of production. To detect these diseases using a traditional method is quite complex. It needs more labor, and the prediction might not be that accurate, either. So, the traditional method is almost impractical for large-scale production. Therefore, automated disease detection can play a significant role here. Automated image-based disease identification methods are emerging to replace the old methods and to help in early diagnosis and avoid human error. These systems could help the farmers to make timely decisions.

Some recent developments of convolutional neural networks (CNNs) have increased the accuracy of leaf disease classification. This model has the capability of automatic feature extraction and advanced pattern detection. In a recent study, Siam et al. presented a lemon leaf dataset that consists of thousands of images classified into healthy and affected classes. They used the DenseNet-121 architecture in their study, and their model achieved an impressive accuracy. This indicates that both dataset diversity and augmentation is required to achieve higher classification performance [1].

Another study was conducted focusing on lemon and orange leaf diseases. They used CNN-extracted features, which are combined with traditional machine learning classifiers. VGG16, VGG19, and ResNet50 are tested in that study for feature extraction. They also applied classifiers such as Random Forest, K-Nearest Neighbor, and Logistic Regression. In this study, they showed the effectiveness of hybrid CNN-ML methods in citrus disease detection [2].

A machine learning-based approach is used in another study for lemon leaf disease classification. They segmented the affected leaf regions using K-means clustering. They extracted texture features with the Gray Level Co-occurrence Matrix (GLCM) and classified them with Support Vector Machines (SVM). This study achieved a good result even with a smaller dataset. They showed that carefully designed feature engineering can still be better than deep CNNs [3].

These studies showed that CNN-based methods and hybrid methods are better than traditional methods of lemon leaf disease detection. But the application of such systems in the real world depends on various things like the availability of diverse datasets, the generalization capability of models, and their adaptability to field conditions where there are variations in lighting, background, and leaf orientation. These are the common challenges to building an effective tool to detect diseases.
Research question:

- To investigate whether hybrid CNN and machine learning approaches improve lemon leaf disease classification.
- To identify the feasibility and challenges of real-time CNN-based lemon leaf disease detection in the field.

## II. Related Works

Lemon plants are vulnerable to various leaf diseases. These diseases cause harm to the production and fruit quality. Recent studies discovered that using deep learning techniques, especially convolutional neural networks (CNNs), can detect diseases in an automatic method. A study by Afrin et al. used applied CNN-extracted features combined with machine learning classifiers like Random Forest and Logistic Regression and achieved 95% accuracy in lemon leaf diseases classification [2]. Another study by Das used XIMR-Net for lemon leaf disease detection and achieved high accuracy, and showed efficiency in automated classification [4].

Traditional machine learning techniques have also proven effective. A study using the technique achieved 94% accuracy by using feature extraction and SVM classification [5]. Besides, hybrid methods combining CNN for feature extraction with classifiers like Random Forest also got a high accuracy in lemon leaf disease detection [6]. CNN-based models alone have shown good performance in precise classification of lemon leaf diseases using architectures like VGG and ResNet [3].

Preprocessing and ensemble approaches also developed the performance of classification [7]. A study integrating image enhancement with a nonlinear fuzzy ranking ensemble showed a good accuracy in lemon leaf disease detection. The study also showed the importance of robust preprocessing for reliable results [8].

S. R. Sahu et el. used a machine-learning-based approach, which is used to classify and detect lemon leaf diseases using K-means clustering for lesion segmentation, GLCM for texture feature extraction, and SVM for classification. The process is well-structured with three stages: segmentation, feature extraction, and classification. The result shows that the method achieves high accuracy. Besides, the processing time is faster than traditional approaches. Use of larger datasets could make the paper stronger and enable detailed performance comparisons with deep-learning models [9].

The proposed segmentation+ GLCM feature extraction and SVM classification approach yields sample-wise classification accuracies ranging roughly from 80% to 90%, which indicates reasonable performance for the dataset used. D. S. Solanki and R. Bhandari's study is based on deep learning-based CNN architectures, specifically GoogleNet, ResNet, and SqueezeNet. These models are used to classify lemon leaf diseases. The dataset contains 609 images (70% training, 30% testing) with and without data augmentation to improve generalization. The result shows that ResNet achieved the highest accuracy of about 97.66%, which indicates strong performance for early disease detection. However, the study also does not compare results with newer architectures or hybrid/ML classifiers. Besides performance under noise, lighting variation or field conditions is not evaluated [10].

Saygılı, A. (2025) [11]. used a hybrid framework which combines transfer-learning-based deep feature extraction with classical machine-learning classifiers. Features are taken from pre-trained CNNs (DenseNet201, ResNet50, MobileNetV2, EfficientNet-B0) and lessens using the mRMR method before being classified with SVM, bagged trees, k-NN, and neural networks under 10-fold stratified validation [12]. DenseNet201 + SVM achieved 94.1% accuracy, which is the best performance. It means that deep features combined with classical classifiers can be highly effective. Further performance gains could be limited because the method is multi-step and depends on pre-trained models rather than end-to-end deep learning.

These studies suggest that the combination of CNN architectures, hybrid classifiers, and image preprocessing techniques helps to achieve better accuracy for lemon leaf detection, which could be used to detect the diseases automatically and help the farmers, thereby [13].

## III. Methodology

Fig. 1 shows the overall methodological workflow of this study. This process starts with the preprocessing and then is split into training, validation, and testing subsets of the dataset. Multiple models were evaluated, and an ensemble of models was used to enhance robustness and reliability. Adversarial training is applied to make the model more stable against noisy inputs. Finally, GRAD-CAM is used to highlight the important regions that influence the model's prediction.

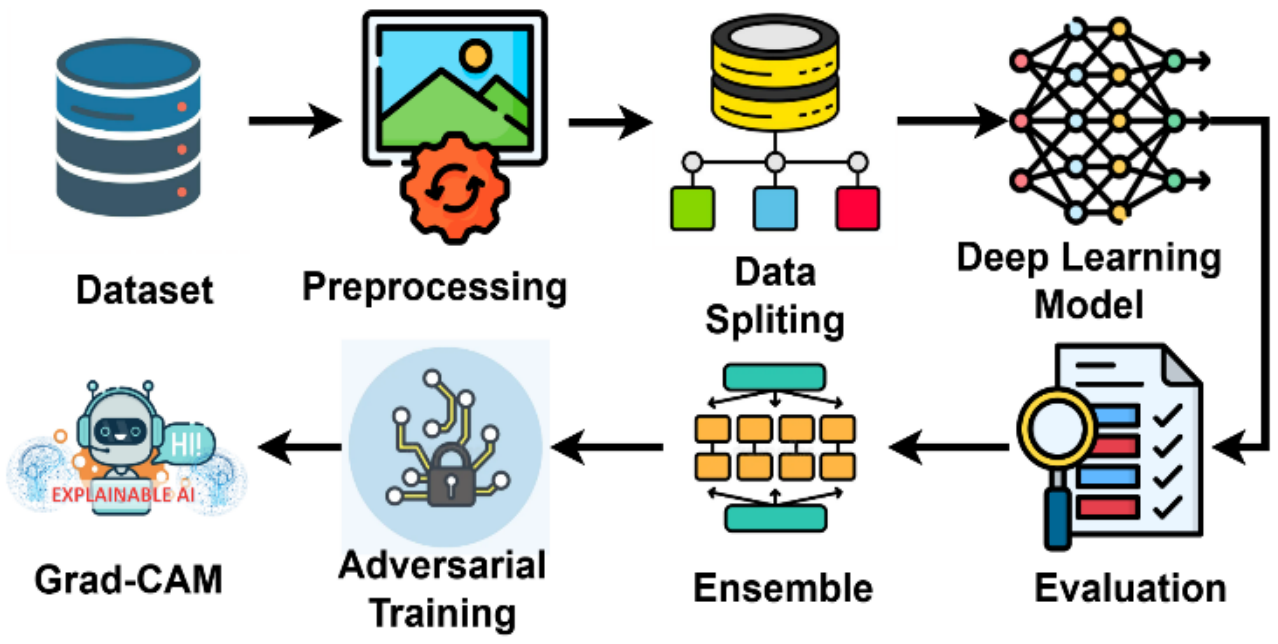

Fig. 1. Methodological framework

### A. Dataset

In this study, the dataset was collected from Kaggle [14] and contains 1354 images. The dataset has 9 classes: Anthracnose (100), Bacterial Blight (105), Citrus Canker (178), Curl Virus(115), Deficiency Leaf(193), Dry Leaf (186), Healthy Leaf (210), Sooty Mould (153), and Spider Mites (114). Fig. 2 shows some examples of images from all classes.

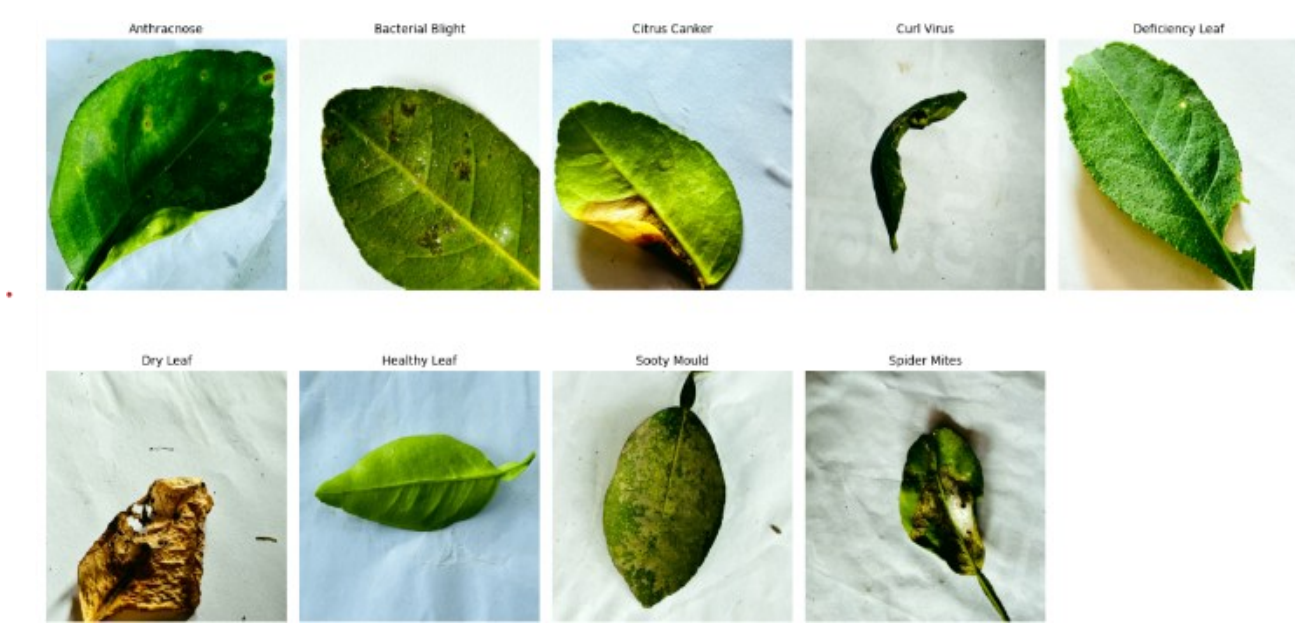

Fig. 2. Sample images from the dataset

### B. Preprocessing

To ensure consistency, all images were resized to 224×224 pixels. The dataset was split into training (70%), test (15%), and validation (15%) sets. This ensures that the evaluation outperforms on unseen samples.

### C. Model Architectures

In this study, we evaluate two different CNN models. These are InceptionV3 and MobileNetv2.

- InceptionV3 outperforms due to its deep layers and efficient convolutions [15]. This enables high accuracy with less computation.
- MobileNetV2 uses a compact architecture with inverted residues and linear bottleneck layers, making it a good trade-off between resource-intensive and high precision models, making it an appropriate choice in resource-constrained applications [16].

These two models were selected. The models are most effective in the agricultural sphere, offering effective and accurate disease detection to maintain crop health. After that, to improve prediction accuracy and enhance robustness, we applied an ensemble based on these two base models [17].

### D. Training Setup

TABLE I demonstrates that the same training method was used in 3 models. The models were trained with 200 epochs. The batch size was set as 32, and the Adam optimizer with a 0.0001 learning rate. Early stopping was used with patience for 10 epochs to mitigate overfitting.

TABLE I. HYPERPARAMETER SETTINGS FOR THE EVALUATED CNN ARCHITECTURES

| Parameter | InceptionV3 | MobilenetV2 | Ensemble (InceptionV3+MobileNetb2) |
|---|---|---|---|
| Batch size | 32 | 32 | 32 |
| Loss function | Categorical cross-entropy | Categorical cross-entropy | Categorical cross-entropy |
| Learning rate | 0.0001 | 0.0001 | 0.0001 |
| Optimizer | Adam | Adam | Adam |
| epochs | 195 | 186 | 169 |
| patience | 10 | 10 | 10 |

## IV. RESULTS

### A. InceptionV3

Fig. 3 illustrates the training and validation accuracy and loss of the InceptionV3 model. Both training and validation accuracy grow rapidly during the early epochs and level off around 99%. Similarly, both training and validation loss drop smoothly to zero with slight variation near epoch 20. Since both curves are close, it indicates that the overall model illustrates strong convergence and well generalization.

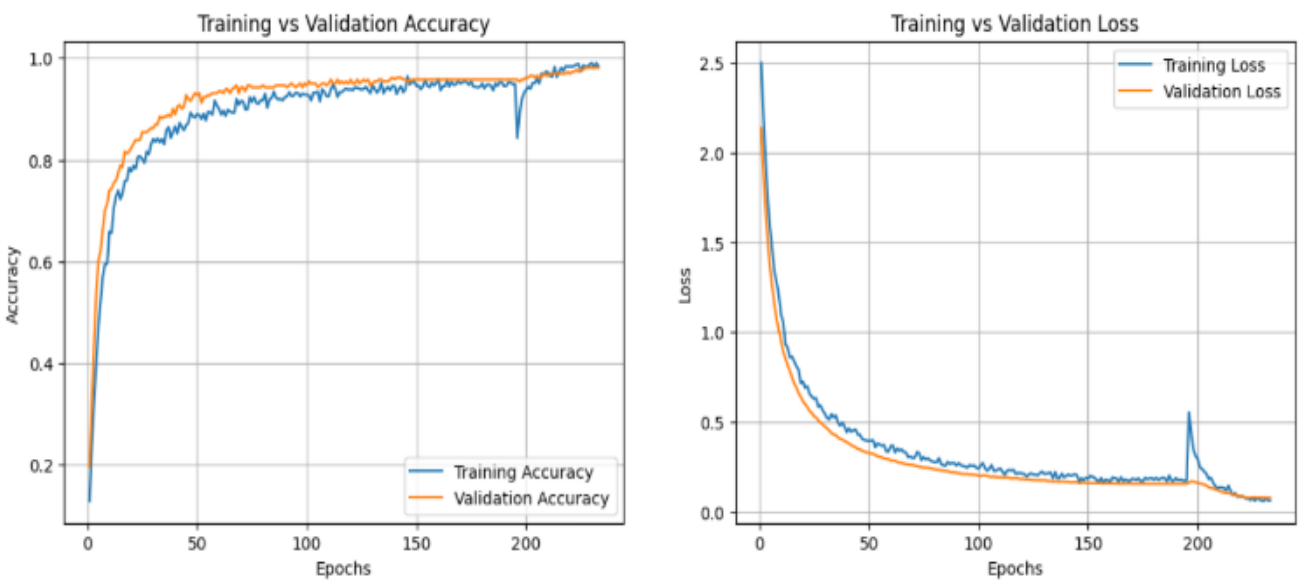

Fig. 3. Accuracy and Loss for training and validation of InceptionV3 architecture

The Fig. 4 represents the confusion matrix for the InceptionV3 model.

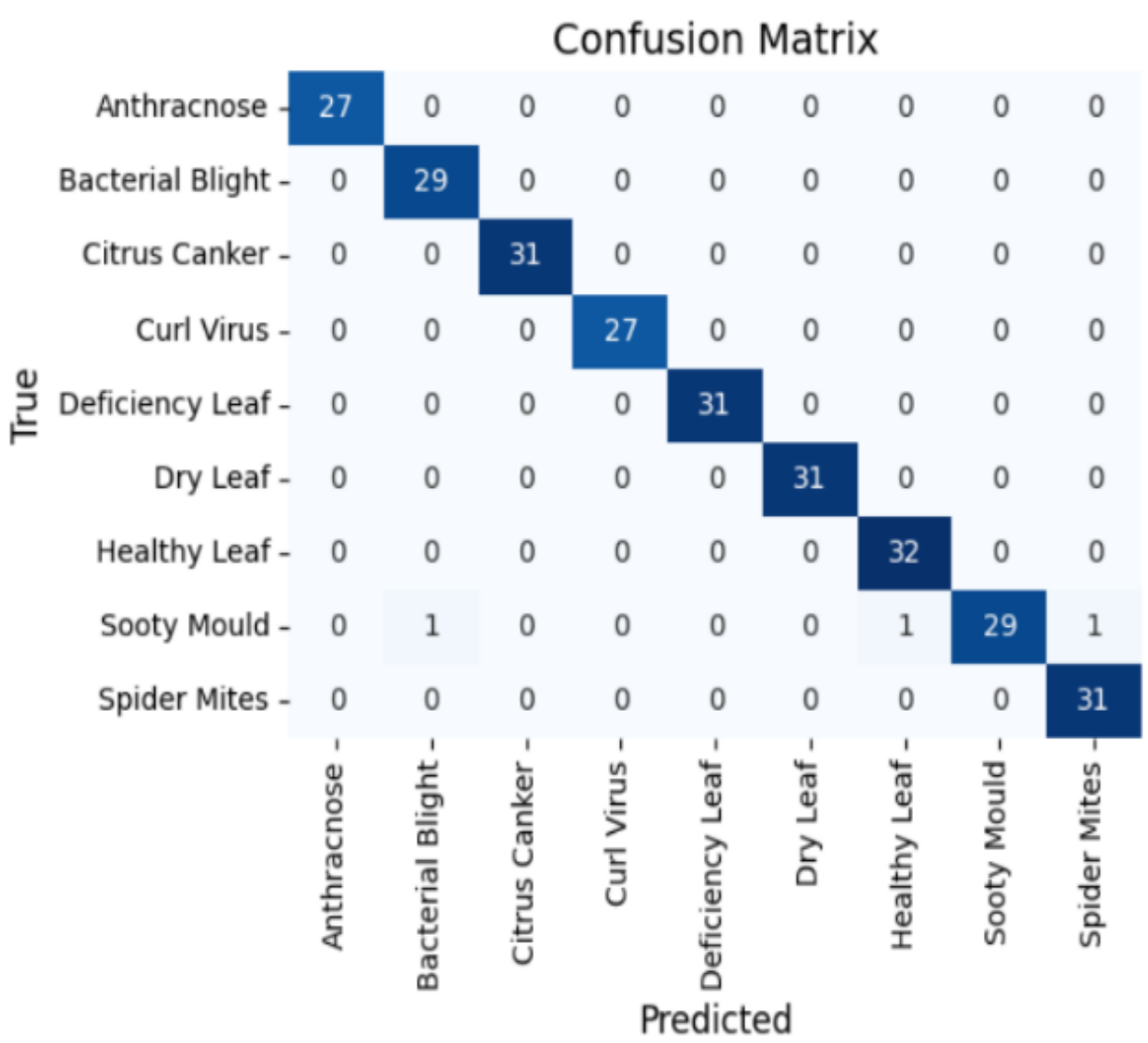

Fig. 4. Confusion Matrix of InceptionV3 architecture

This reveals the model correctly predicted all the classes with high accuracy, along with some minor misclassification in the Sooty Mould class. This Sooty Mould class was misclassified as Bacterial Blight, and one as Spider Mites. Nevertheless, since the errors are negligible in comparison to the overall predictions, the model achieves excellent performance in general. This validates that the model is not only most effective, but also reliable, in the differentiation of a wide range of conditions of the leaf.

### B. MobileNetV2

Fig. 5 shows the overall results of the training and validation curves of MobileNetV2 exhibiting strong convergence and validation accuracy consistently exceeding 97%.

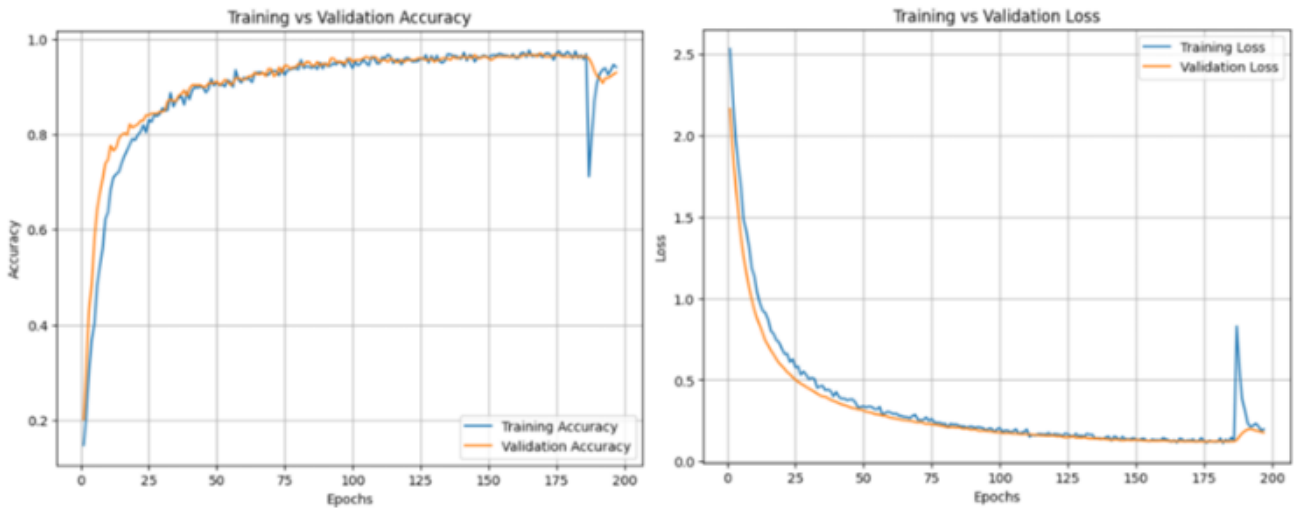

Fig. 5. Training and validation accuracy and loss curve of MobileNetV2 archirecture

Fig. 6 shows the confusion matrix of MobileNetV2 model. Most of the classes correctly classified. There are few misclassifications occur some Citrus Canker predicted as Deficiency leaf and Healthy leaf confused with Sooty Mould or Anthracnose. *Overall, the model has good performance with classifications.*

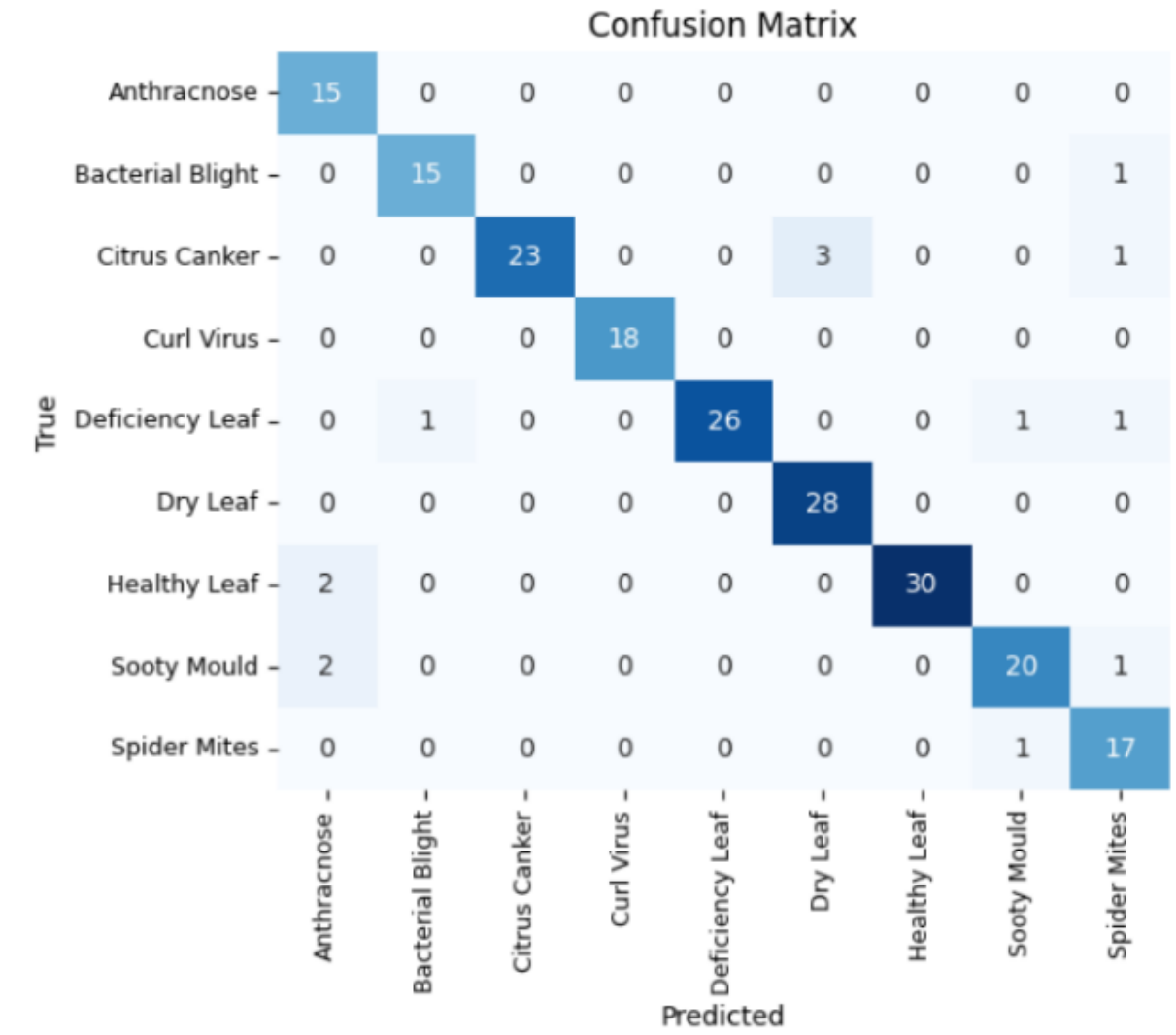

Fig. 6. Confusion Matrix of MobileNetV2 architecture

### C. Ensemble

Fig. 7 presents the proposed ensemble model architecture. Two pretrained CNNs, InceptionV3 and MobileNetV2 are applied to extract high level features from the input images.

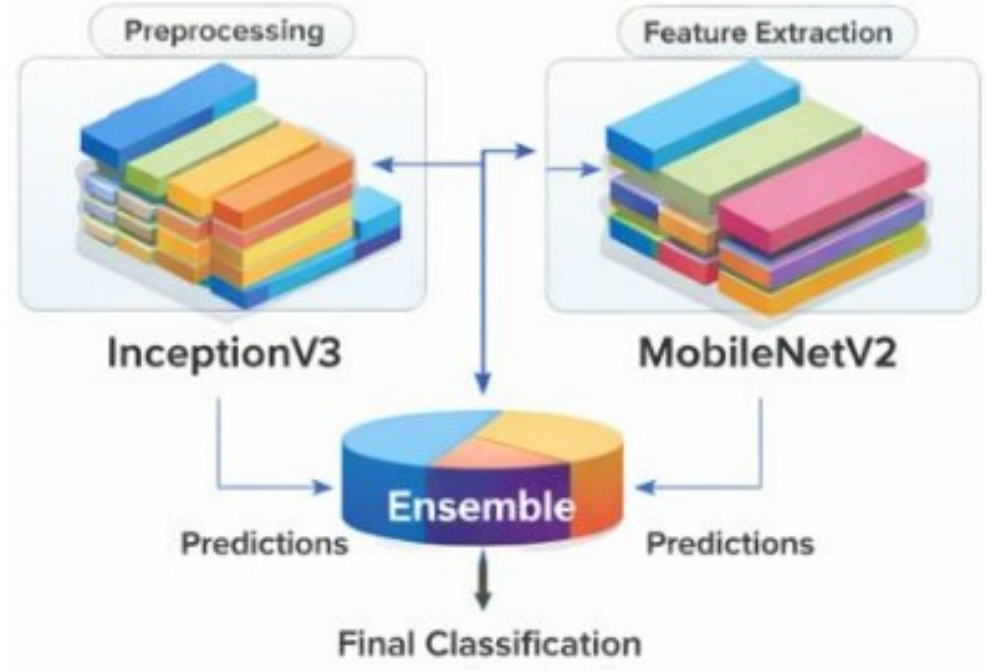

Fig. 7. Ensemble Model Architecture

The predictions from both models are combined in an ensemble fusion strategy to obtain a single final prediction.

Fig. 8 illustrates the ensemble model training and validation accuracy and loss curves. The ensemble model achieved around 100% validation accuracy with steady growth and training accuracy closely followed by only a tiny temporary dip. The validation loss stayed lower than the training loss, which proves that the ensemble model has strong generalization ability.

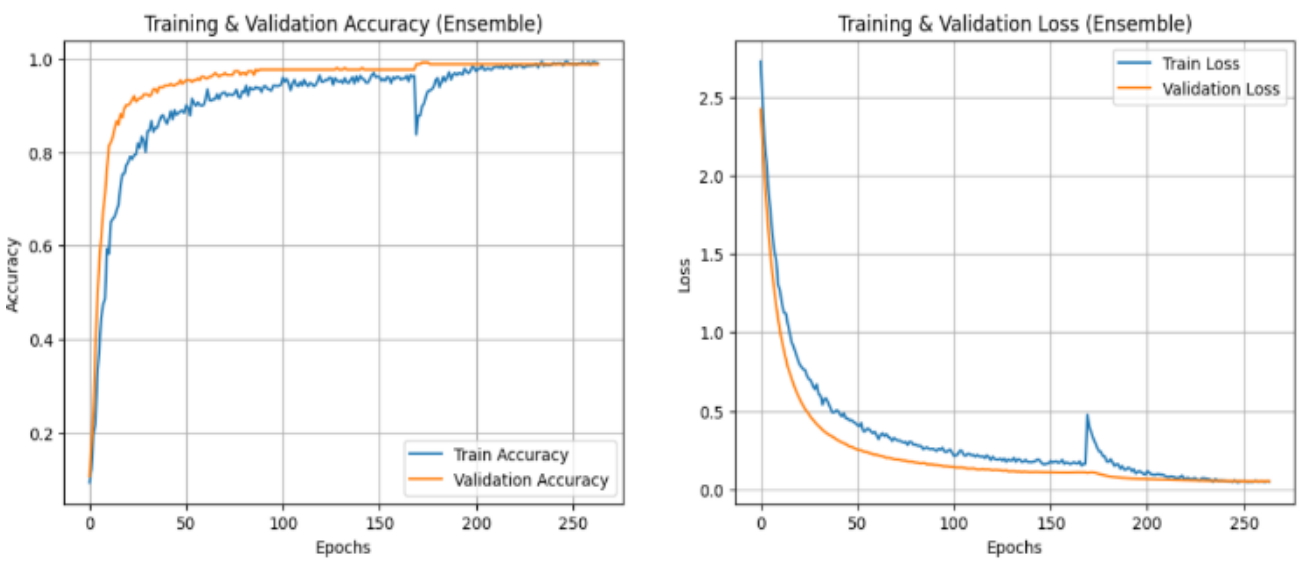

Fig. 8. Accuracy and Loss for training and validation of the Ensemble architecture

The confusion matrix in Fig. 9 shows that the ensemble model achieved almost perfect classification over all lemon leaf diseases. This ensemble model correctly predicts all the classes with 100% accuracy. Only one Curl Virus was misclassified as Anthracnose, and one Sooty Mould was classified as a healthy leaf.

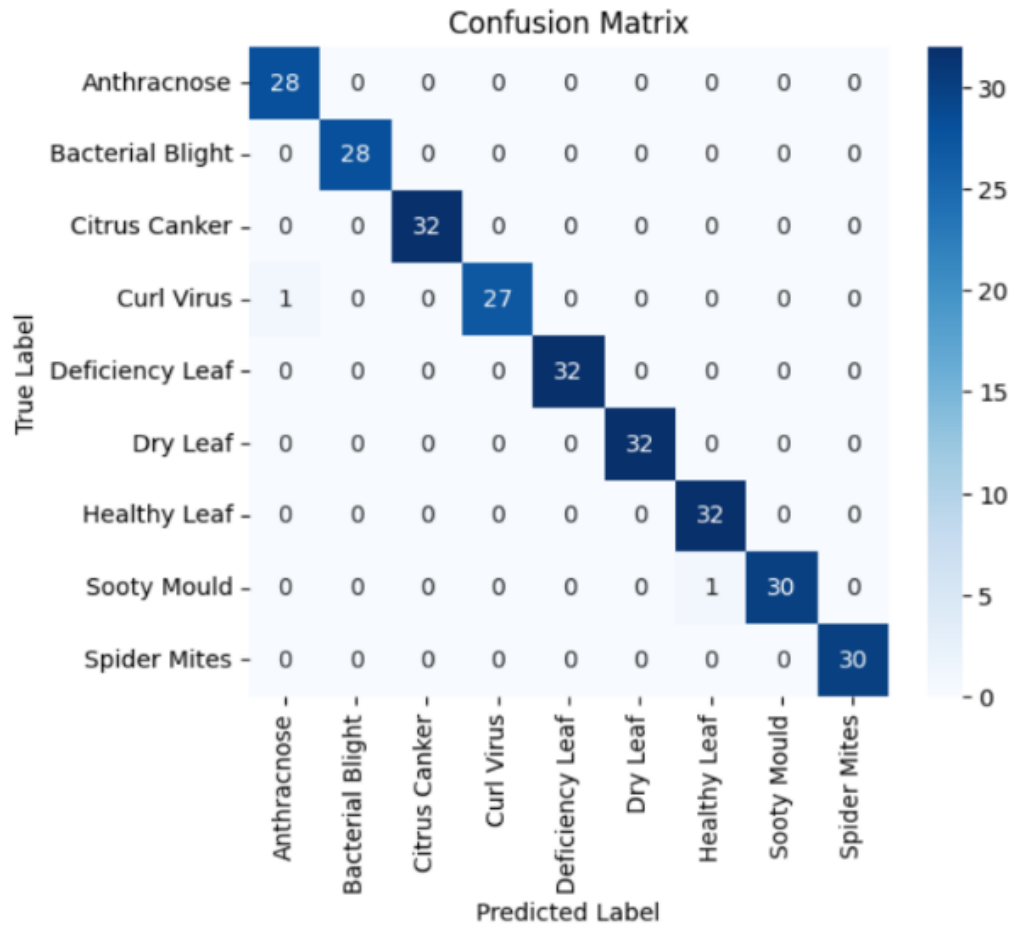

Fig. 9. Confusion Matrix of Ensemble architecture

### D. Adversarial Robustness

Adversarial training is applied to enhance the robustness of the model. Here, our proposed Ensemble model achieved 84.70% peak accuracy at ε= 0.1. Table II shows the stability of the Ensemble model, with analyses promising for noisy images.

TABLE II. ADVERSARIAL ROBUSTNESS EVALUATION OF XCEPTION UNDER VARYING PERTURBATION LEVELS ( $\varepsilon$ )

| Epsilon Value (ε) | Validation Loss | Validation Accuracy | Optimal Epochs |
|---|---|---|---|
| 0 | 0.0346 | 0.9962 | 199 |
| 0.1 | 0.6456 | 0.8470 | 159 |
| 0.12 | 0.6231 | 0.8246 | 186 |
| 0.14 | 0.8292 | 0.7276 | 80 |
| 0.16 | 1.0972 | 0.6492 | 62 |
| 0.18 | 1.4104 | 0.4888 | 40 |
| 0.2 | 1.6741 | 0.4514 | 58 |

## V. EXPLAINABLE AI

### A. Grad-CAM Visualizations

The Fig. 10 shows the GRAD-CAM visualization of the ensemble model. This visualization highlights [18] the infected regions of the lemon leaf. This indicates that the ensemble model successfully classified spider mites' disease of the lemon leaf with 99.52% confidence.

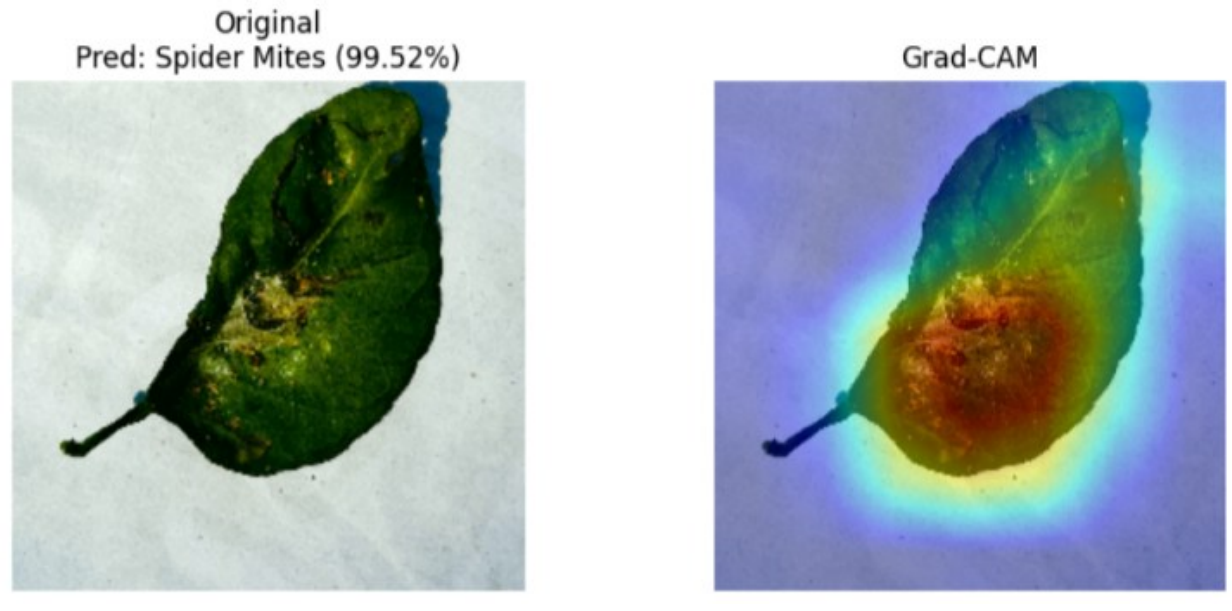

Fig. 10. Lemon leaf disease identification using Grad-CAM

## VI. PROTOTYPE WEB APPLICATION

Fig. 11 highlights the output of the lemon leaf disease recognition system.

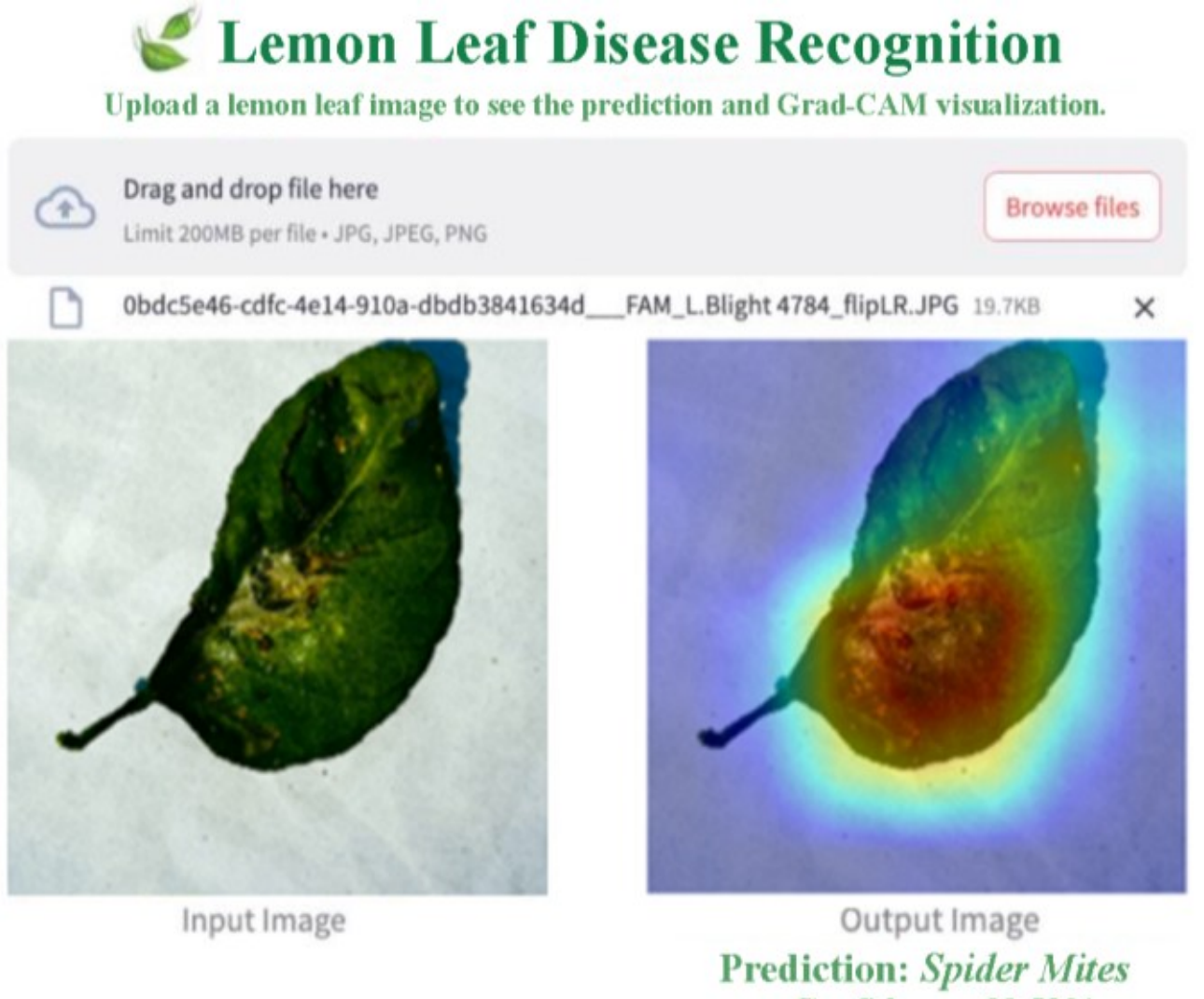

Fig. 11. LeafLife web application for lemon leaf disease classification

The original image is shown on the left, while the output image presents the Grad-CAM visualization. This highlights the most informative diseased region for prediction [19]. The model correctly predicts the leaf class as Spider mites with a confidence score of 99.52%.

## VII. COMPARISON AND DISCUSSION

### A. Evaluation of CNN Architectures

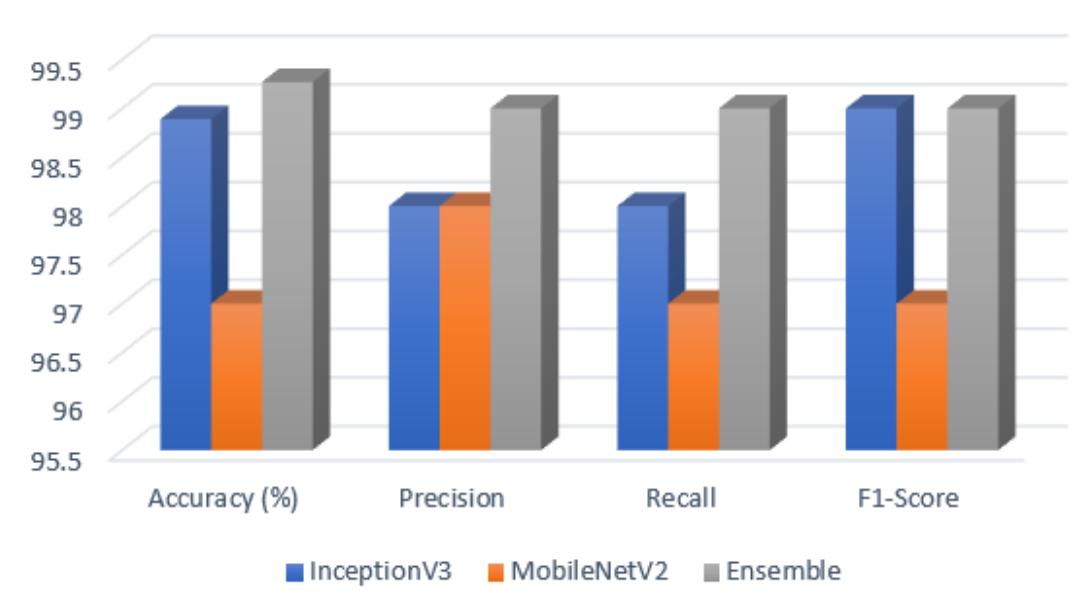

Fig. 12. Performance Comparison of Evaluated CNN Architectures

Fig. 12 represents a comparative analysis of the performance metrics for InceptionV3, MobileNetV2, and the Ensemble model. The ensemble model achieved the highest 99.27% test accuracy.

### *B. Benchmarking Against Prior Work*

Table III shows three studies on lemon and citrus leaf disease detection. Khandoker et. al. [2] used deep learning models like VGG16, VGG19, and ResNet50 with traditional classifiers such as Random Forest, Naive Bayes, KNN, and Logistic Regression, and achieved 95% accuracy. A K M et al. used the DenseNet-121 model, which achieved 98.56 % accuracy [1]. Besides, Kaur et al. [3] tested architectures like VGG19, AlexNet, and Xception, and their model achieved 96.51% accuracy [3].

TABLE III. COMPARISON OF LEAFLIFE WITH PREVIOUS WORKS ON LEMON LEAF DISEASE CLASSIFICATION

| Author | Method | Accuracy (%) |
|---|---|---|
| Khandoker et. al. [2] | VGG19 | 95 |
| Kaur et. al. [3] | Ensemble (VGG19, AlexNet, and Xception) | 96.51 |
| A K M et. al. [1] | DenseNet201 | 98.56 |
| Proposed Model | InceptionV3 | 98.89 |
| | MobileNetv2 | 97 |
| | Ensemble | 99.27 |

## VIII. CONCLUSION

Healthy lemon crops are crucial to boost production and quality. This study is designed to mitigate the challenges to enhance lemon disease detection using deep learning models.

InceptionV3 (98.89%) and MobileNetV2 (97.45%) are ensembles in this study. The ensemble model achieved 99.27% accuracy for lemon disease detection. Adversarial training was integrated into the learning process to increase the model's resilience to noise and unseen distortions. For improving the interpretability of the predictions, Grad-CAM is used. Besides the high accuracy, there are certain limitations in the study, such as a small dataset, inconsistency in image quality, suboptimal preprocessing, and the classes are also imbalanced. In future work, the use of large, high-quality, and balanced datasets could be more useful. Besides, preprocessing steps can be improved to enhance detection accuracy and robustness.

## REFERENCES

[1] Siam, A. K. M. F. K., Bishshash, P., Nirob, M. A. S., Mamun, S. B., Assaduzzaman, M., & Noori, S. R. H. (2025). *A comprehensive image dataset for the identification of lemon leaf diseases and computer vision applications*. *Data in Brief, 58*, 111244. https://doi.org/10.1016/j.dib.2024.111244s

[2] Arifin, K. N., Rupa, S. A., Anwar, M. M., & Jahan, I. (2024). Lemon and orange disease classification using CNN-extracted features and machine learning classifier. arXiv. https://doi.org/10.48550/arXiv.2408.14206

[3] Kaur, B., Gupta, S. K., Janarthan, M., Alsekait, D. M., & AbdElminaam, D. S. (2025). *Precision diagnosis of citrus leaf diseases using image enhancement and nonlinear fuzzy ranking ensemble approach NLFuRBe*. *Scientific Reports, 15*, 32296. https://doi.org/10.1038/s41598-025-16923-4

[4] Sudha, V., Hemalatha, U., & Shankar, S. G. (2024). *Lemon leaf disease detection using machine learning*. *International Journal of Computer Science and Engineering, 11*. https://doi.org/10.14445/23488387/IJCSE-V11I1P101

[5] Das, I., Mim, M. M. H., Kafli, M. A. H., & Utsha, C. B. (2024). *XIMR-Net: A robust deep learning model for automated lemon leaf disease classification*. *ResearchGate*. https://doi.org/10.1109/ECAI65401.2025.11095565

[6] Kalim, H., Chug, A., & Singh, A. P. (2022). *Citrus leaf disease detection using hybrid CNN-RF model*. *ResearchGate*. https://doi.org/10.1109/AIST55798.2022.10065093

[7] Yadav, P. K., Burks, T., Frederick, Q., Qin, J., Kim, M., & Ritenour, M. A. (2022). *Citrus disease detection using convolution neural network generated features and Softmax classifier on hyperspectral image data*. *Frontiers in Plant Science, 13*. https://doi.org/10.3389/fpls.2022.1043712

[8] Ghosh, S. (2025). Healthy harvests: A comparative look at guava disease classification using InceptionV3. 16th International IEEE Conference on Computing, Communication and Networking Technologies (ICCCNT). IEEE. https://doi.org/10.48550/arXiv.2602.10967

[9] Sahu, S. R., Jena, S., & Panda, S. (2025). A machine learning approach for classification of lemon leaf diseases. In Proceedings of CoCoLe 2022 (pp. 254–265). Springer. https://doi.org/10.1007/978-3-031-21750-0_22

[10] Solanki, D. S., & Bhandari, R. (2022). Lemon leaf disease detection and classification at early stage using deep learning models. NeuroQuantology, 20(22), 3455–3468. https://doi.org/10.48047/nq.2022.20.22.NQ10345

[11] Saygılı, A. (2025). Enhancing lemon leaf disease detection: A hybrid approach combining deep learning feature extraction and mRMR-optimized SVM classification. Applied Sciences, 15(20), 10988. https://doi.org/10.3390/app152010988

[12] Shikdar, O. F. et al. Evaluating Tea Leaf Quality with Deep Learning: Insights from Occlusion Sensitivity Analysis. International Conference on Computing and Communication Networks (ICCCNet-2025). Springer.

[13] Noor, M. T., Alam, B. M. S., Kibria, G., Moontaha, S., & Ahammed, F. (2024). Deep learning for green growth: Custom CNN uncovers cucumber leaf disease secrets. *International Journal for Multidisciplinary Research*, *6*(6), Article 32327. https://doi.org/10.36948/ijfmr.2024.v06i06.32327

[14] Shaheen, M. (2024). Lemon leaf disease dataset (LLDD) [Data set]. Kaggle.https://www.kaggle.com/datasets/mahmoudshaheen1134/lemon-leaf-disease-dataset-lldd

[15] Hossain, F. (2025). Toward reliable and explainable nail disease classification: Leveraging adversarial training and Grad-CAM visualization. 16th International IEEE Conference on Computing, Communication and Networking Technologies (ICCCNT). IEEE. https://doi.org/10.48550/arXiv.2602.04820

[16] Shikdar, O. F. (2025). Enhancing tea leaf disease recognition with attention mechanisms and Grad-CAM visualization. In Proceedings of the International Conference on Computing and Communication Networks (ICCCNet-2025). Springer. https://doi.org/10.48550/arXiv.2512.17987

[17] Alam, B. M. S. (2025). Explainable Deep Learning for Hog Plum Leaf Disease Diagnosis: Leveraging the Elegance of the Depth of Xception. International Conference on Green Energy, Computing and Sustainable Technology' (GECOST 2025). IEEE. https://doi.org/10.1109/GECOST66002.2025.11324532

[18] Ahammed, F., Shikdar, O. F., Alam, B. M. S., Jahan, S., Kibria, G., & Ali, N. Y. (2025). Classifying nutritional deficiencies in coffee leaf using transfer learning and gradient-weighted class activation mapping (Grad-CAM) visualization. *Proceedings of the 2025 International Conference on Computing and Communication Technologies (ICCCT)*, 1-5. https://doi.org/10.1109/ICCCT63501.2025.11019090

[19] Kibria, G. (2025). MediVision: An Explainable and Robust Deep Learning Framework for Knee Osteoarthritis Grading. 2025 IEEE International Conference on Data and Software Engineering (ICoDSE). IEEE. https://doi.org/10.1109/ICoDSE68111.2025.11351659